# Meta Transfer Learning for Facial Emotion Recognition

Dung Nguyen*, Kien Nguyen*, Sridha Sridharan*, Iman Abbasnejad*, David Dean* and Clinton Fookes*
*Queensland University of Technology
2 George St, Brisbane, QLD 4000, Australia
Email: {d.nguyentien, k.nguyenthanh, s.sridharan, i.abbasnejad, c.fookes}@qut.edu.au, ddean@ieee.org

***Abstract*—The use of deep learning techniques for automatic facial expression recognition has recently attracted great interest but developed models are still unable to generalize well due to the lack of large emotion datasets for deep learning. To overcome this problem, in this paper, we propose utilizing a novel transfer learning approach relying on PathNet and investigate how knowledge can be accumulated within a given dataset and how the knowledge captured from one emotion dataset can be transferred into another in order to improve the overall performance. To evaluate the robustness of our system, we have conducted various sets of experiments on two emotion datasets: SAVEE and eNTERFACE. The experimental results demonstrate that our proposed system leads to improvement in performance of emotion recognition and performs significantly better than the recent state-of-the-art schemes adopting fine-tuning/pre-trained approaches.**

## I. Introduction

Emotions of humans manifest in their facial expressions, voice, gestures, and posture. An accurate emotion recognition system using one or a combination of these modalities would be useful in various applications including surveillance, medical, robotics, human computer interaction, affective computing, and automobile safety [1]. Researchers in this area have focused mainly in the area of facial expression recognition to build reliable emotion recognition systems. This is still a challenging problem since very subtle emotional changes manifested in the facial expression could go undetected [1]. Recently several approaches based on deep learning techniques have contributed to progress in this area [1]–[10]. However a major problem which is still hindering progress is the lack of large annotated facial expression databases for deep learning.

To overcome the lack of large facial expression datasets, many recent studies [11]–[13] have exploited transfer learning approaches based on fine-tuning/pre-trained models to the facial emotion recognition task, where they have shown that the models pre-trained on large-scale datasets could be transferred to the facial emotion recognition task. In spite of some success, such fine-tuning based transfer learning approaches still have some shortcomings as pointed out by Rusu *et al* [14]. In particular, transfer learning across multiple tasks was inappropriate. If we wish to leverage on the knowledge acquired over a sequence of experiences, this begs the question which model should be used to initialize subsequent models? This might require not only a learning approach with the capability of supporting transfer learning without catastrophic forgetting, but also prior knowledge of task similarity [14]. While fine-tuning might allow recovery of expert performance in the target domain, it is often a destructive process due to discarding prior learned information [14]. Moreover, to completely remember all previous tasks, each model might be copied before fine-tuning, however this did not remove the problem of choosing an appropriate initialization [14]. Finally, transferring learned parameters between inconsistent architectures could be challenging [15]. Such limitations of the current transfer learning approaches have been a setback in realising the full potential of transfer learning.

To address these drawbacks of transfer learning techniques based on fine-tuning approach, Gideon *et al.* [15] have proposed the use of progressive networks [14] which are able to potentially support transferring knowledge across sequences of tasks. Gideon *et al.* [15] have successfully transferred learning between three paralinguistic tasks: emotion, speaker, and gender recognition, with an emphasis on emotion detection as the target domain without the catastrophic forgetting effect. Their system outperformed the recent modern approaches utilizing fine-tuning/pre-trained model and deep leaning models without transfer learning [15].

Nonetheless, an unavoidable limitation of this approach [14] is that it is computationally intensive since the number of new networks keeps on growing according to the demand for the increasing number of new tasks which need to be learned [16], [17].

To alleviate the aforementioned downsides, Fernando *et al.* [18] have recently proposed PathNets as an alternative novel learning algorithm for transfer learning. PathNets is designed as a neural network in which instead of creating a new column of networks and combining the output of prior columns with the current one when transferring learned knowledge into new tasks (as in [14]), agents (i.e. pathways through the network) are embedded to discover which parts of the network to re-use for new tasks [18]. Agents also hold an accountability for determining which the subset of parameters to use and update. Pathways through the neural network for replication and mutation are selected by a tournament selection genetic algorithm [19] during learning. The parameters along an optimal path evolved on the source task are fixed and a new population of paths for the destination task is re-evolved. Such manner of transfer learning enables the destination task to be learned faster than learning from scratch or fine-tuning [18]. The authors have demonstrated the success on several supervised learning classification problems.

Motivated by the success of PathNets in other applications, in this paper we explore utilizing PathNets for facial emotion recognition. We focus on investigating how effective PathNets will be in transferring knowledge captured from the SAVEE dataset [20] into the eNTERFACE dataset [21], and how emotional knowledge can be accumulated in one dataset itself (i.e. SAVEE, eNTERFACE) to improve overall performance. The contributions of our paper are the following:

1) We introduce a novel application of transfer learning approach utilizing PathNets in dealing with the insufficient emotion data, and catastrophic forgetting issue for the facial expression recognition task.
2) We confirm, through experimental results, that our system has significant potential to accurately detect emotions, and demonstrate its substantial success in transferring learned knowledge between different emotion datasets.

The remainder of this paper is organized as follows: Section 2 describes related research; Section 3 presents our proposed system; Section 4 reports our experimental results; and Section 5 concludes the paper.

## II. Related Research

A number of studies using deep learning techniques have recently been proposed for the facial expression task. These will be discussed in the following sub-sections.

### A. Facial Expression Recognition Using Deep Learning

Hamester *et al.* [2] proposed multi-channel convolutional neural networks (a standard CNN and a channel using pre-trained parameters, which were obtained by a convolutional auto-encoder (CAE)) for facial expression recognition. Teixeira *et al.* [3] have confirmed that convolutional neural networks significantly outperform conventional architectures such as LBP + SVM and SIFT + SVM. To address the non-rigidity and motion involved in facial expression, Liu *et al.* [4] have recently proposed the use of improved optical flow [5] to generate optical flow of facial expressions, which was subsequently fed into the stacked sparse auto-encoder (SAE) to extract high-level features.

In other studies, Neagoe *et al.* [6] have investigated CNNs and deep belief networks (DBN) separately to resolve a subject's independent/dependent facial expression recognition. Jung *et al.* [7] have proposed three models: deep temporal appearance networks (DTANs), deep temporal geometry network (DTGN), and deep temporal appearence-geometry network (DTAGN). While the first model applied CNN to extract temporal appearance features from image sequences, the second one adopted DNN to detect geometrical movement of information from the facial landmark points. These two models were combined in DTAGN to capture these features simultaneously.

Although the above proposed systems have shown competitive performance, the use of spatio-temporal features for facial expression recognition task has not been effectively addressed. For more effective use of the spatio-temporal features for emotion recognition, researchers have proposed the use of Convolutional Three Dimensional Network (C3D). Nguyen *et al.* [1] have proposed the learning of spatio-temporal features with C3Ds from audio and video. Fan *et al.* [8] also exploited C3Ds to model appearance and motion of video simultaneously. Abbasnejad *et al.* [9] trained C3D on generated synthetic data. They achieved the competitive recognition accuracy. Ghasemi *et al.* [10] also adopted C3D to extract spatio-temporal features, which were subsequently fed into the LSTM network that was able to evolve the dynamic duration of these features representing emotion progresses from the neutral state to the apex state. Another approach, which was quite in line with Ghasemis's idea, has been proposed in [22], in which 3D Inception-ResNet layers followed by LSTM unit were designed. By such design, their system successfully extracted the spatial relations within facial images as well as the temporal relations between a contiguous image sequence.

### B. Meta Transfer Learning

Research into facial expression recognition has been hindered by the lack of large databases for learning [11]–[13]. To address the lack of large emotion dataset, the fine-tuning/pre-trained model has been recently widely investigated for facial expression recognition task [11]–[13]. The CNN architectures were pre-trained using generic ImageNet dataset and fine-tuned on facial expression datasets [12], [13]. To learn visual segment and audio features, Zhang *et al* [11] used pre-trained the CNN models in a 3DCNN on large-scale image and video classification datasets, and then fine-tuned them on emotion recognition tasks. However, as discussed earlier, all of these fine-tuning approach based methodologies [11]–[13] still have the drawbacks previously alluded to such as discarding previously learned information which were detailed by Rusu *et al* [14]. In another research, Gideon *et al.* [15] have introduced a learning algorithm using the progressive networks proposed by Andrei *et al.* [14] to effectively transfer knowledge captured from one emotion dataset into another. Although they have handled somewhat successfully the above mentioned limitations, their expensive computation, which kept on increasing when adding new tasks to be learned, makes them less applicable in the implementation of real-time emotion recognition.

## III. Proposed Methodology

In this paper, we propose the use of a novel application of a meta transfer learning approach exploiting PathNets in tackling facial emotion detection task. Our architecture is illustrated in Fig. 1. We initially exploit a Viola Jones-based algorithm [1] to extract all face regions from both SAVEE [20] and eNTERFACE [21] datasets, which are subsequently fed into PathNets to classify a final emotion score. To the best of our knowledge, the use of PathNets have not been investigated to address the dearth of suitable emotion databases for the development of facial expression recognition.

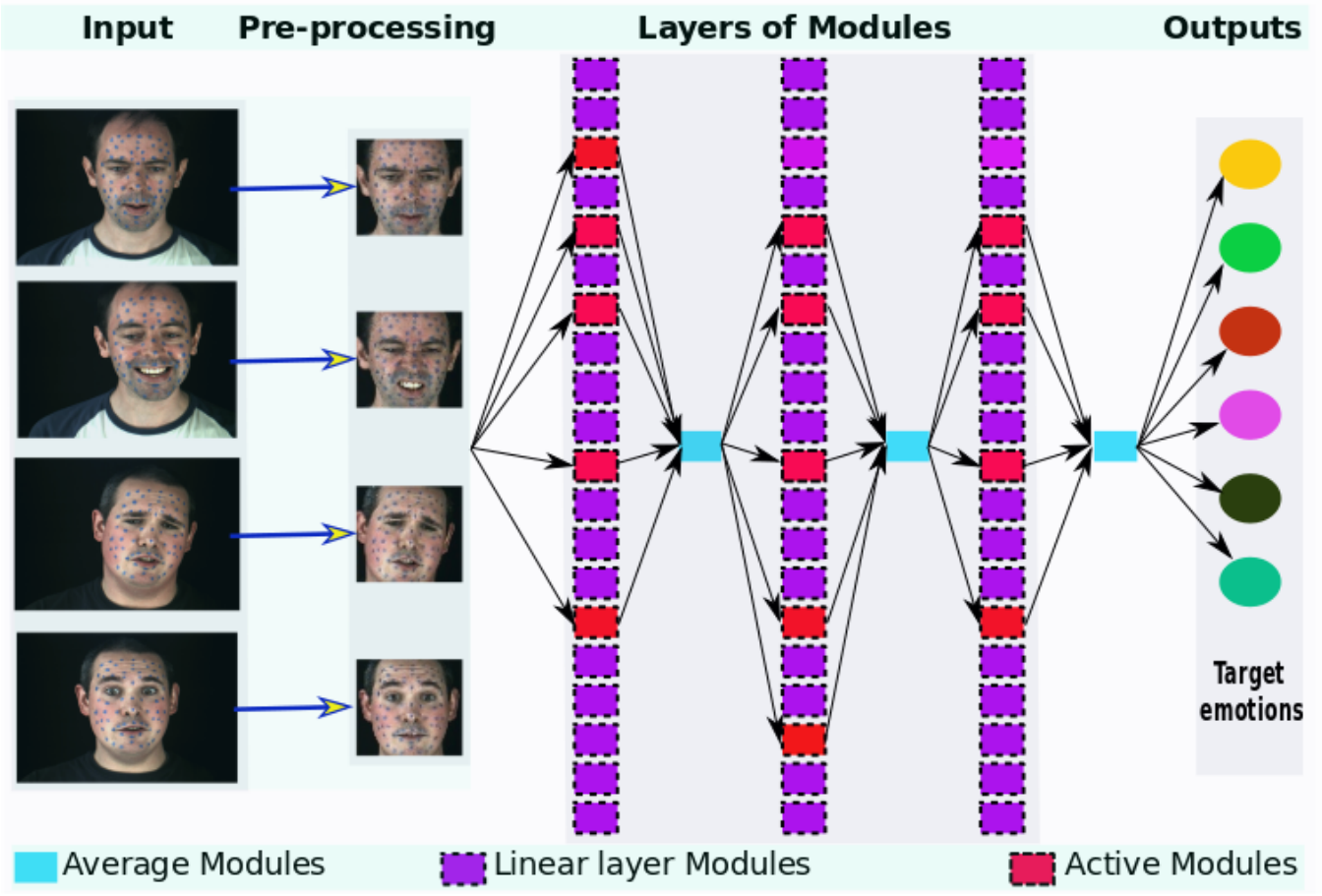

Fig. 1. This illustrates our PathNet architecture which is designed similar to PathNet [18] used to conduct all sets of experiments on CIFAR [23] SVHN [24] consisting of 3 layers. Each layer includes 20 modules of 20 neurons in each, and each module is followed by a rectified linear unit. Between layers the feature maps are averaged before being fed into the modules of the next layer. A maximum of 5 modules per layer are typically allowed to be presented in a pathway.

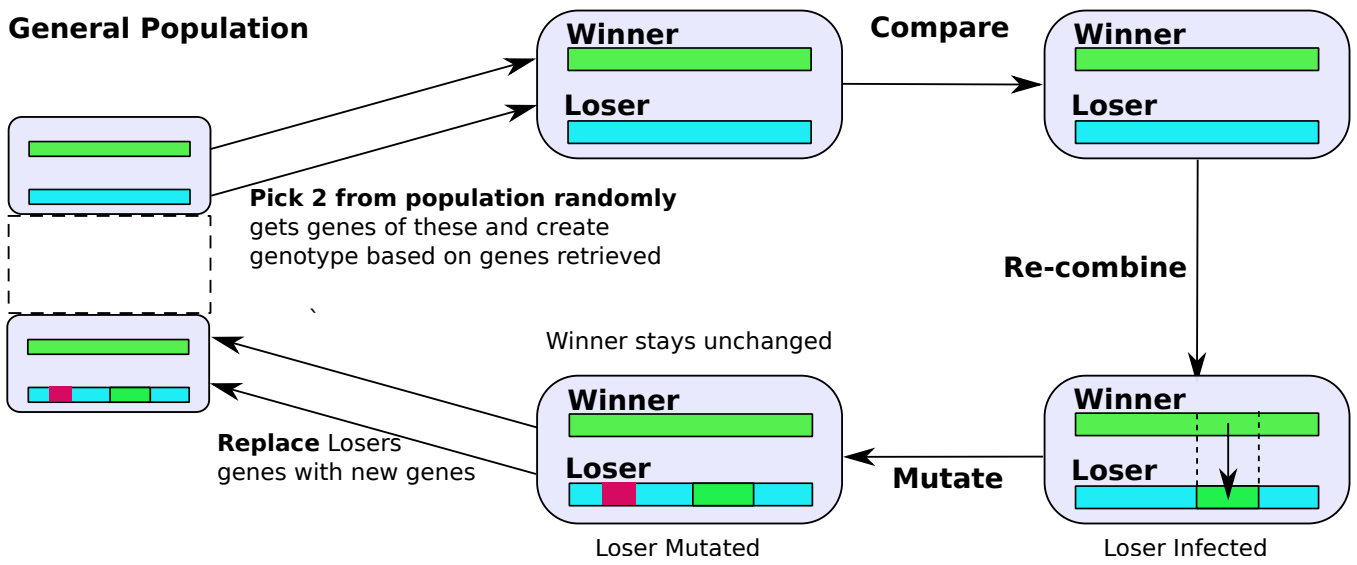

Fig. 2. The genotypes of the population are viewed as a pool of strings. One single cycle of the Microbial GA is operated by initially **randomly picking** two, and subsequently **compare** their fitnesses to determine **Winner**, **Loser**, and finally **recombine** where some proportion of Winner's genetic material infects the **Loser**, before **mutating** the revised version of **Loser** [19].

### *A. Pre-processing*

All frames are initially extracted from visual signal for further steps. Since such extracted frames still contain considerable redundant information for emotion detection, we extract only the face regions using the simple algorithm [1] as follows:

1) All bounding boxes containing face regions in each frame were extracted employing the Viola-Jones algorithm [25] and a face region was then detected.
2) In some cases where the Viola Jones algorithm detected no faces, or more than 1 face, the location of the previously detected face region was used.

By applying this algorithm, we have successfully extracted all face regions from all frames in both datasets (SAVEE and eNTERFACE) as input into PathNets (see **Pre-processing** column in Fig. 1. as an illustration of some input samples)

### *B. PathNets*

Our PathNet architecture and its settings relies on PathNets [18] which was used to conduct all sets of experiments on CIFAR [23] and SVHN [24]. The following sections will provide in detail its architecture and explain further how to train our system, how to transfer learned emotional knowledge.

*1) PathNet architecture:* Our PathNets includes a number of layers (L = 3), a number of modules (M = 20) per layer of 20 neurons in each. Each module itself functions as a neural network consisting of linear units, and followed by a transfer function (rectified linear units adopted). For each layer the outputs of the modules of this layer are averaged before being fed into the active modules of the subsequent layer. A module is active if it is shown in the path genotype and currently validated (shown in Fig. 1.). A maximum of 5 distinct modules per layer are typically allowed in a pathway. The final layer is not shared for each task which is being learned [18].

*2) Pathway Evolution and Transfer Learning Method:* Either eNTERFACE or SAVEE (source task) is trained during a fixed period of time with a goal of finding an optimal pathway by adopting a binary tournament selection algorithm [19] (see Fig. 2.) which is responsible for eliminating bad configurations and mutating the good ones, and subsequently training them further. This pathway is then fixed (i.e. its parameters no longer permitted to modify). All other parameters, which are not represented in such best fit path, are reinitialized and trained/evolved on the same dataset (eNTERFACE or SAVEE) or on other dataset (destination task). Such transferring method allows destination task to be learned faster than learning from scratch or after fine-tuning. The performance measurement of our system is the accuracy of recognition obtained after such fixed traning time. Evidence to prove positive transfer in these cases is provided by the destination task showing a better final emotion recognition accuracy in comparison with learning from scratch [18].

## IV. Experiments & Results

We evaluated our proposed methodology on the eNTERFACE [21] and SAVEE [20] datasets. All our experiments are conducted applying the 6-fold-cross-validation protocol, and implemented using the Tensorflow deep learning framework.

### *A. Dataset Analysis*

The eNTERFACE dataset [21] consists of 1166 video in which the number of men and women are 902 (77%) and 264 (23%) respectively expressing anger, disgust, fear, happiness, sadness, and surprise.

The SAVEE dataset [20] is an audio-visual dataset which was recorded from higher degree researchers (aged from 27 to 31 years) at the University of Surrey, and four native male British speakers expressing seven discrete emotions (anger, disgust, fear, happiness, sadness, surprise, and neutral). The dataset comprises of 120 utterances per speaker, resulting in a total of 480 sentences [20].

In this paper, we address only six basic emotions for consistency between two datasets, and extract video stream from both datasets for conducting all experiments.

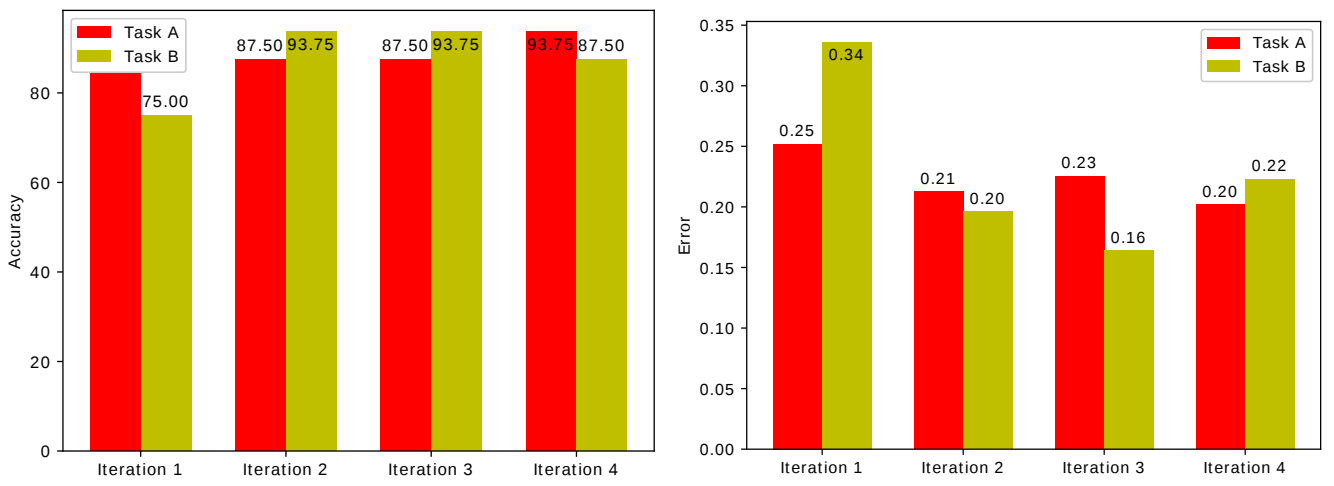

(a) Accuracy on SAVEE (b) Traning loss on SAVEE

Fig. 3. Comparison the performance of our system between 4 iterations when accumulating learned emotional knowledge on one dataset itself, i.e. source task (Task A) and destination task (Task B) on the same emotion SAVEE dataset

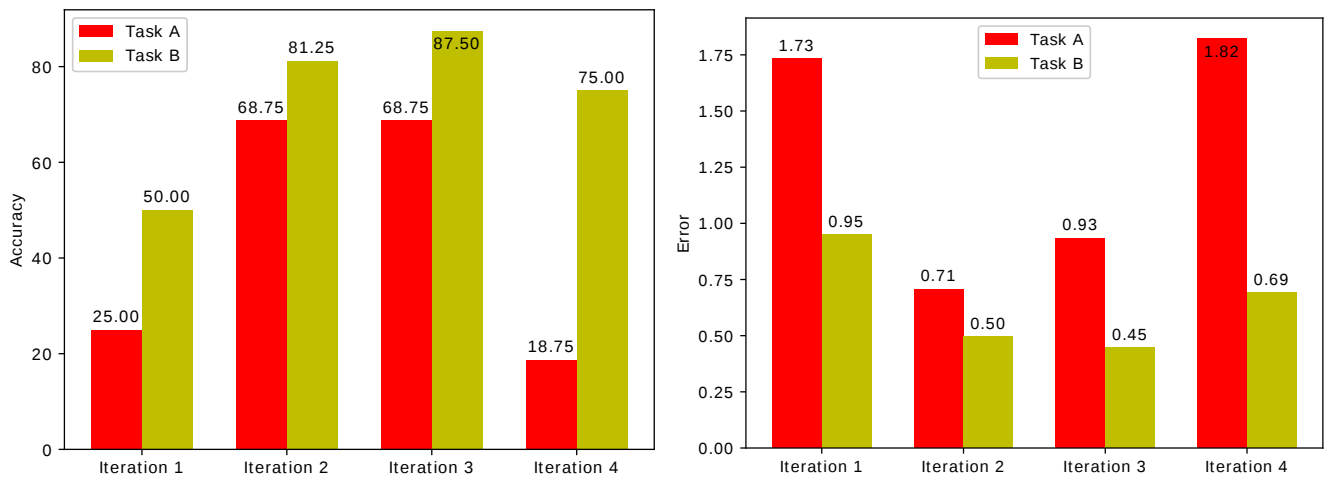

(a) Accuracy on eNTERFACE (b) Training loss on eNTERFACE

Fig. 5. Comparison the performance of our system between 4 iterations when accumulating learned emotional knowledge on one dataset itself, i.e. source task (Task A) and destination task (Task B) on the same emotion eNTERFACE dataset

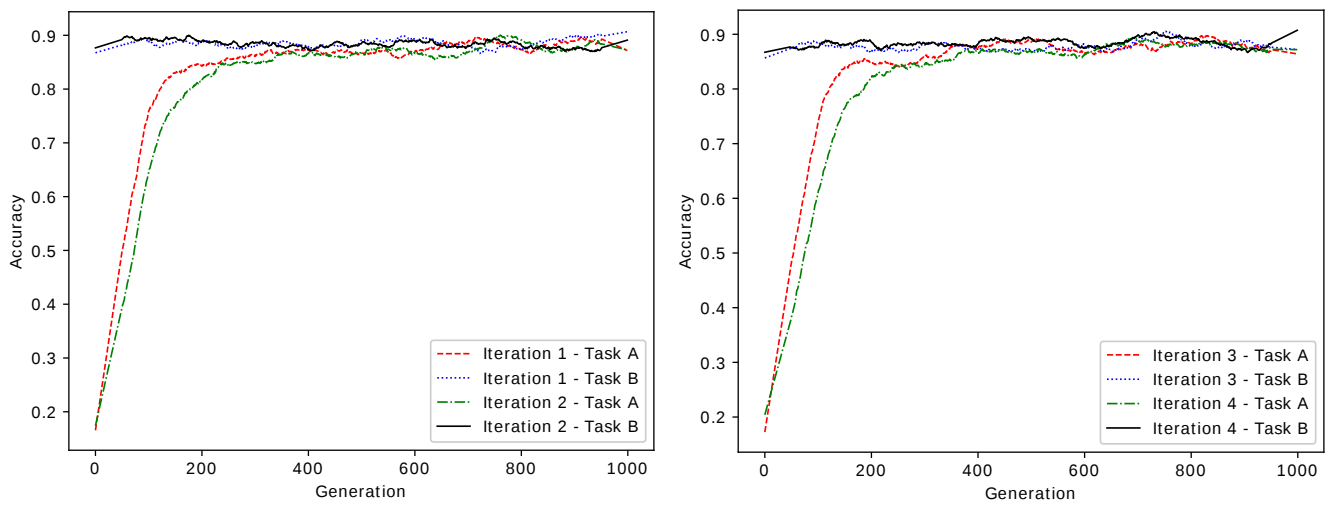

(a) Accuracy for iteration 1, 2 (b) Accuracy for iteration 3, 4

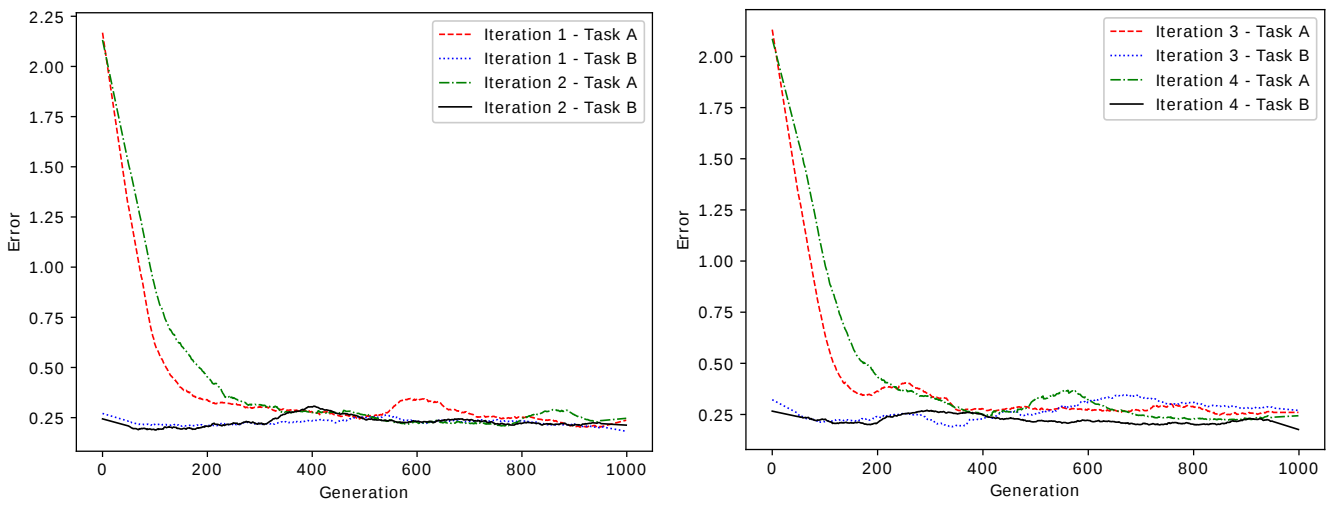

(c) Loss for iteration 1, 2 (d) Loss for iteration 3, 4

Fig. 4. We illustrate the different learning curves of our system (accuracy and loss shown in (a), (b), and (c), (d) respectively) for 4 iterations during training when transferring learned emotional knowledge on SAVEE dataset itself, i.e. source task - Task A (SAVEE), destination task - Task B (SAVEE)

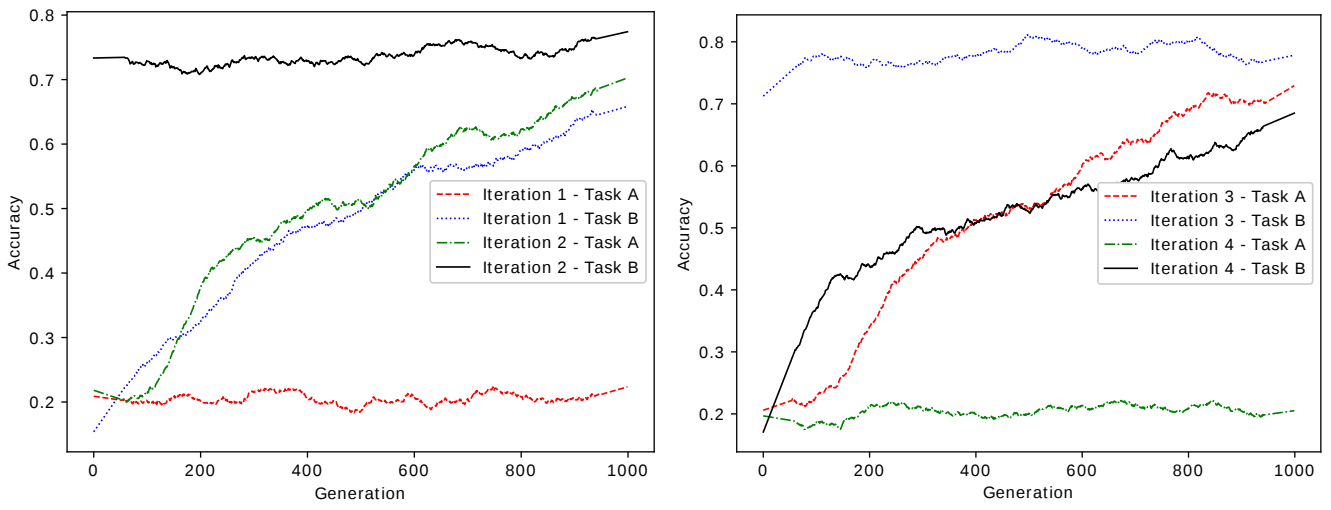

(a) Accuracy for iteration 1, 2 (b) Accuracy for iteration 3, 4

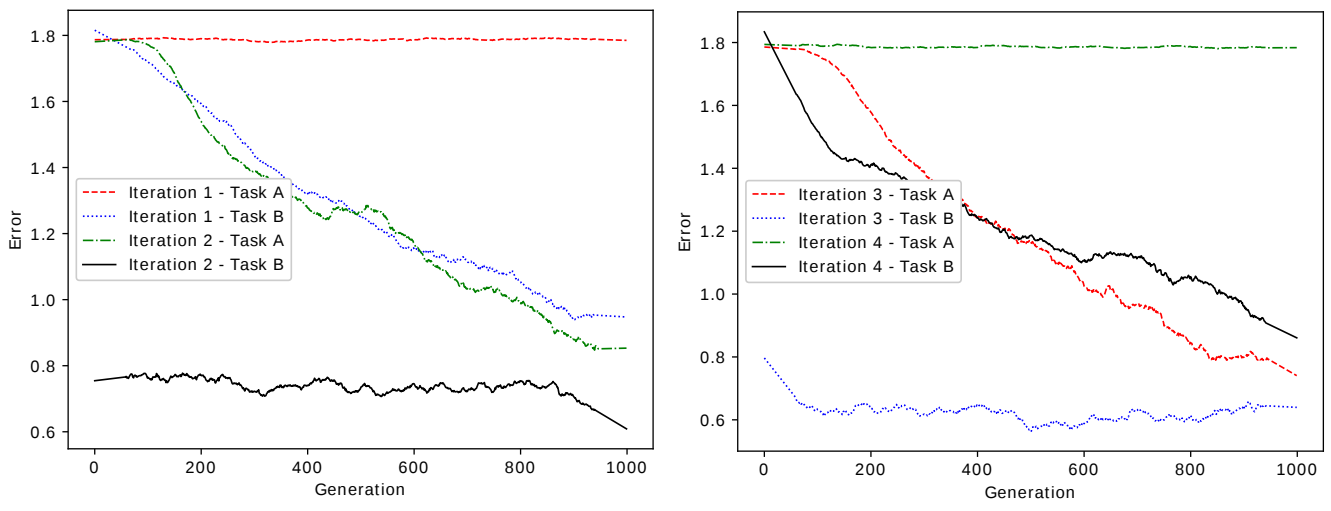

(c) Loss for iteration 1, 2 (d) Loss for iteration 3, 4

Fig. 6. We illustrate the different learning curves of our system (accuracy and loss shown in (a), (b), and (c), (d) respectively) for 4 iterations during training when transferring learned emotional knowledge on eNTERFACE dataset itself, i.e. source task - Task A (eNTERFACE), destination task - Task B (eNTERFACE).

### B. *Using off-the-shelf CNNs as Baseline Systems*

In this set of experiments, we use some pre-trained models (Cifar, Inceptions, Mobilenet, Resnet, and Vgg) and fine-tune them on two datasets: eNTERFACE, and SAVEE. The performance of these models is used as baseline systems to compare with our proposed one. The training loss rates are illustrated in Table I. When fine-tuning Inception, Mobilenet, and Resnet models on eNTERFACE, these models face the under-fitting problem.

### C. *PathNets*

We conduct three sets of experiments using PathNets to examine how emotion recognition accuracy improves when transferring learned knowledge between datasets. In order to investigate further the effectiveness of our network, we run each set of experiments with four iterations and compare the performance between them. In the first two sets, PathNets is trained on SAVEE/eNTERFACE as source tasks and then evolved on SAVEE/eNTERFACE as destination tasks respectively. In the third set, we train PathNets on SAVEE as the source task, and then evolve on eNTERFACE as the destination task.

*1) PathNet Settings:* Our PathNet architecture (i.e. L = 3 layers, M = 20 liner units per layer of 20 neurons each followed by rectified linear units, average function used to activate units between two layers, and a maximum of 5 of those units per layer included in a pathway) is trained for 1000 generations on source and destination task. At the beginning of each task, a population of 20 of these genotypes is randomly created and evolved on the destination task. In each generation, two paths are randomly selected for validation. To evaluate one path, our network is trained with stochastic gradient descent with learning rate 0.02 on 50 mini-batches of size 16 and one pathway is trained for 50 epochs. The fitness of such pathway is the rate of correct samples on the set of training during that period of training time. When completing the calculation of the fitness of two pathways, the pathway with the smaller fitness is replaced by the pathway with the greater fitness and mutated. So, the network between tasks is modified as follows: the parameters presented in the best fit path which is evolved on the source task, are fixed, and the rest of parameters are randomly reinitialized when evolved on the destination task [18].

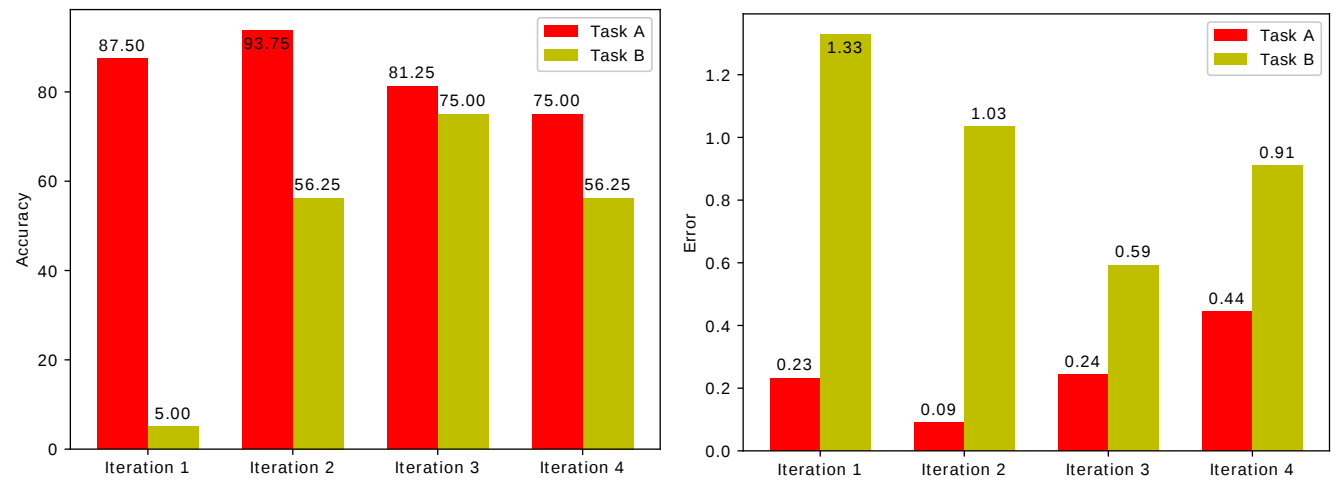

(a) Accuracy (b) Loss

Fig. 7. Comparison performance of our system (accuracy and loss shown in (a), and (b) respectively) between 4 iterations when transferring learned emotional knowledge from source task (Task A): SAVEE dataset into destination task (Task B): eNTERFACE dataset

| METHODOLOGY | Loss (SAVEE) | Loss (eNTERFACE) |
|---|---|---|
| Cifar [26] | 1.773 | 1.3263 |
| Inception_v1 [27] | 1.812 | - |
| Inception_v2 [28] | 1.8 | - |
| Inception_v3 [28] | 1.777 | - |
| Inception_v4 [29] | 1.773 | - |
| Mobilenet_v1 [30] | 1.791 | - |
| Mobilenet_v2 [30] | 1.796 | - |
| Mobilenet_v3 [30] | 1.798 | - |
| Mobilenet_v4 [30] | 1.775 | - |
| Resnet_v1_50 [31] | 1.812 | - |
| Resnet_v1_101 [31] | 1.827 | - |
| Resnet_v1_152 [31] | 1.886 | - |
| Resnet_v1_200 [31] | 1.824 | - |
| Resnet_v2_50 [32] | 1.787 | - |
| Resnet_v2_101 [32] | 1.772 | - |
| Resnet_v2_152 [32] | 1.813 | - |
| Resnet_v2_200 [32] | 1.823 | - |
| Vgg_19 [33] | 1.792 | 1.56 |
| Our proposed system | 0.44 | 0.91 |
| Our proposed system | **0.22** | **0.69** |

TABLE I
THE TRAINING LOSS OF DIFFERENT DEEP LEARNING NETWORKS WHICH ARE FINE-TUNED FROM DIFFERENT PRE-TRAINED CNN MODELS ON TWO DATASETS: SAVEE AND ENTERFACE

*2) Experimental results:* We report and compare a series of experimental results of our system on three afore-mentioned sets of experiments. In the first two sets of experiments, the performance (training accuracy and training loss) of our system on SAVEE and eNTERFACE are illustrated in Fig. 3. and Fig. 5. respectively. As indicated by these figures, our system exhibits a significant improvement in its performance (recognition accuracy and training loss) on destination task when trained/evolved and transfer learned emotional knowledge from source task (i.e. the system is trained from scratch). Our system achieves best recognition accuracy (93.75% and 87.5% on SAVEE and eNTERFACE respectively) on destination task at iteration 3, which are (6.25% and 18.75%) higher than those achieved by learning from scratch (see Fig. 3a. and Fig. 5a.). Our system also obtains a low training loss rates on the same task and the same iteration (0.16 and 0.45 on SAVEE and eNTERFACE respectively), which are (0.7 and 0.48) lower than those achieved by training from scratch (see Fig. 3b. and 5b.) and substantial lower than those of all our baseline systems utilizing fine-tuning approach (shown in Table. I.). The different learning curves of our system on these datasets are illustrated in Fig. 4. and Fig. 6.

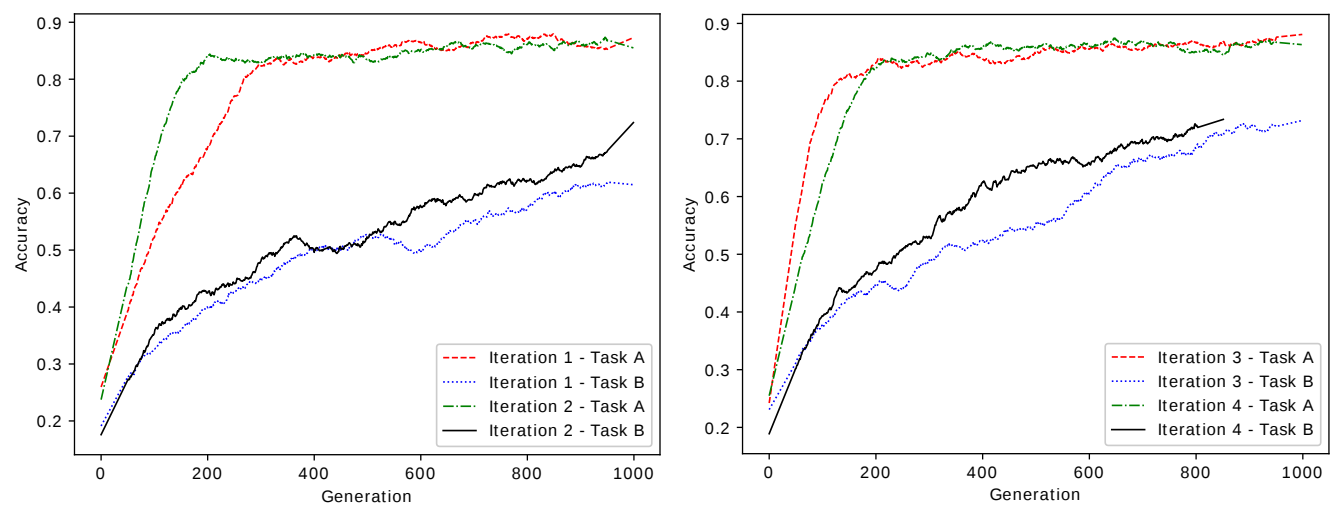

(a) Accuracy for iteration 1, 2 (b) Accuracy for iteration 3, 4

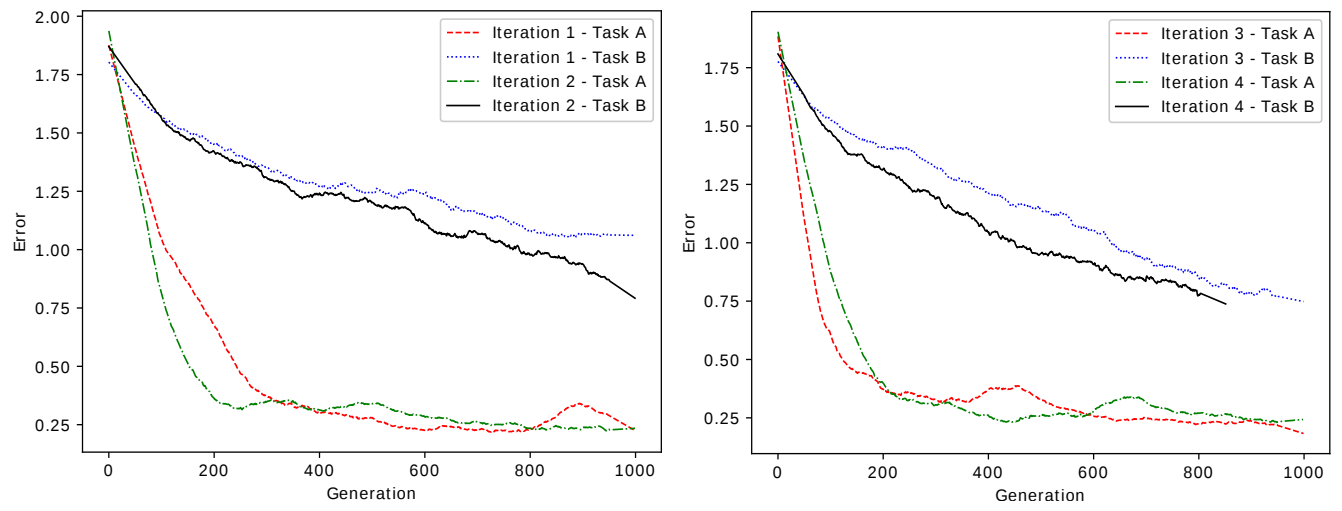

(c) Loss for iteration 1, 2 (d) Loss for iteration 3, 4

Fig. 8. We illustrate the different learning curves of our system (accuracy and loss shown in (a), (b), and (c), (d) respectively) for 4 iterations during training when transferring learned emotional knowledge from source task (Task A): SAVEE dataset into destination task (Task B): eNTERFACE dataset.

To further demonstrate the robustness of our system, we conduct an additional set of experiments using PathNets with the goal of transferring learned emotional knowledge captured from SAVEE into eNTERFACE. As illustrated in Fig. 7a., at the first iteration, in contrast to the high recognition accuracy (87.5%) of our system achieved on SAVEE when learning from scratch with PathNet, our system achieves only 5% of recognition accuracy when eNTERFACE is learned as the destination task with our system. However, this accuracy dramatically increases to 75% at iteration 3. Similarly, the training loss rate of our system on eNTERFACE trained with PathNet as destination task considerably decreases from 1.33 at iteration 1 to 0.59 at iteration 3 which still much lower than those of our baselines (see Table. I) and very competitive with that obtained by learning with PathNet on the same dataset (eNTERFACE). The different learning curves of our system on this set of experiments are illustrated in Fig. 8.

When compared with our baseline systems, our system achieves lowest training loss rates (0.22 and 0.69) on SAVEE dataset and eNTERFACE dataset respectively which are 0.355 and 0.6363 lower than the best performance of our pre-trained models on SAVEE and eNTERFACE respectively (illustrated in Table I). While pre-trained models: Inception, Mobilenet, Resnet face under-fitting issue when fine-tuning them on eNTERFACE dataset, our proposed system performs very well on this dataset.

## V. Conclusion

One of the major challenges in developing a robust emotion recognition system is to achieve well generalized models with limited data. To address this problem, researchers have investigated the use of transfer learning but the proposed techniques suffer from issues such as discarding prior learned information. In this paper, by utilizing the PathNet as an alternative transfer learning mechanism, we have addressed the problem of insufficient data for deep learning. With our approach of transfer learning, we have successfully accumulated learned emotion knowledge in one dataset and successfully transferred the knowledge captured from one dataset into another, consequently resulting in an improved emotion recognition rate. The experimental results validated on SAVEE and eNTERFACE shows that our proposed system outperforms the state-of-the-art transfer learning exploiting fine-tuning methodology by a significant margin. Our future research will focus on the use of PathNets in transferring knowledge learned from multiple tasks into facial expression domain task, as well as accumulating knowledge learned from multiple emotion datasets into one another.

## VI. Acknowledgement

This research was supported by an Australian Research Council (ARC) Discovery grant No: DP140100793.

## References

[1] D. Nguyen, K. Nguyen, S. Sridharan, A. Ghasemi, D. Dean, and C. Fookes, "Deep spatio-temporal features for multimodal emotion recognition," in *WACV*, March 2017, pp. 1215–1223.

[2] D. Hamester, P. Barros, and S. Wermter, "Face expression recognition with a 2-channel convolutional neural network," in *IJCNN*, July 2015, pp. 1–8.

[3] A. Teixeira Lopes, E. de Aguiar, and T. Oliveira-Santos, "A facial expression recognition system using convolutional networks," in *2015 28th SIBGRAPI*, Aug 2015, pp. 273–280.

[4] Y. Liu, X. Hou, J. Chen, C. Yang, G. Su, and W. Dou, "Facial expression recognition and generation using sparse autoencoder," in *Smart Computing (SMARTCOMP), 2014 International Conference on*, Nov 2014, pp. 125–130.

[5] D. Sun, S. Roth, and M. Black, "Secrets of optical flow estimation and their principles," in *CVPR*, June 2010, pp. 2432–2439.

[6] V. E. Neagoe, B. Andrei-Petru, N. Sebe, and P. Robitu, "A deep learning approach for subject independent emotion recognition from facial expressions," *Recent Advances in Image, Audio and Signal Processing*, pp. 93–98, 2013.

[7] H. Jung, S. Lee, S. Park, I. Lee, C. Ahn, and J. Kim, "Deep temporal appearance-geometry network for facial expression recognition," *CoRR*, vol. abs/1503.01532, 2015.

[8] Y. Fan, X. Lu, D. Li, and Y. Liu, "Video-based emotion recognition using CNN-RNN and C3D hybrid networks," in *Proceedings of the 18th ACM International Conference on Multimodal Interaction*, ser. ICMI 2016. New York, NY, USA: ACM, 2016, pp. 445–450.

[9] I. Abbasnejad, S. Sridharan, D. Nguyen, S. Denman, C. Fookes, and S. Lucey, "Using synthetic data to improve facial expression analysis with 3D convolutional networks," in *ICCV*, Oct 2017.

[10] A. Ghasemi, M. Baktashmotlagh, S. Denman, S. Sridharan, D. N. Tien, and C. Fookes, "Deep discovery of facial motions using a shallow embedding layer," in *2017 ICIP*. IEEE SigPort, 2017.

[11] S. Zhang, S. Zhang, T. Huang, W. Gao, and Q. Tian, "Learning affective features with a hybrid deep model for audio-visual emotion recognition," *IEEE Transactions on Circuits and Systems for Video Technology*, vol. PP, no. 99, pp. 1–1, 2017.

[12] H.-W. Ng, V. D. Nguyen, V. Vonikakis, and S. Winkler, "Deep learning for emotion recognition on small datasets using transfer learning," in *ICMI*. New York, NY, USA: ACM, 2015, pp. 443–449.

[13] H. Kaya, F. Grpnar, and A. A. Salah, "Video-based emotion recognition in the wild using deep transfer learning and score fusion," *Image and Vision Computing*, vol. 65, no. Supplement C, pp. 66 – 75, 2017, multimodal Sentiment Analysis and Mining in the Wild Image and Vision Computing.

[14] A. A. Rusu, N. C. Rabinowitz, G. Desjardins, H. Soyer, J. Kirkpatrick, K. Kavukcuoglu, R. Pascanu, and R. Hadsell, "Progressive neural networks," *CoRR*, vol. abs/1606.04671, 2016.

[15] J. Gideon, S. Khorram, Z. Aldeneh, D. Dimitriadis, and E. M. Provost, "Progressive neural networks for transfer learning in emotion recognition," *CoRR*, vol. abs/1706.03256, 2017.

[16] S.-W. Lee, J.-H. Kim, J. Jun, J.-W. Ha, and B.-T. Zhang, "Overcoming catastrophic forgetting by incremental moment matching," in *2017 NIPS*, I. Guyon, U. V. Luxburg, S. Bengio, H. Wallach, R. Fergus, S. Vishwanathan, and R. Garnett, Eds. Curran Associates, Inc., 2017, pp. 4655–4665.

[17] A. Mallya and S. Lazebnik, "Packnet: Adding multiple tasks to a single network by iterative pruning," *CoRR*, vol. abs/1711.05769, 2017.

[18] C. Fernando, D. Banarse, C. Blundell, Y. Zwols, D. Ha, A. A. Rusu, A. Pritzel, and D. Wierstra, "Pathnet: Evolution channels gradient descent in super neural networks," *CoRR*, vol. abs/1701.08734, 2017.

[19] I. Harvey, "The microbial genetic algorithm," in *2009 ECRL*, ser. ECAL'09. Berlin, Heidelberg: Springer-Verlag, 2011, pp. 126–133.

[20] S. Haq and P. Jackson, "Speaker-dependent audio-visual emotion recognition," in *Proc. Int. Conf. on Auditory-Visual Speech Processing (AVSP'08), Norwich, UK*, Sept. 2009.

[21] O. Martin, I. Kotsia, B. Macq, and I. Pitas, "The enterface' 05 audio-visual emotion database," in *Data Engineering Workshops, 2006. Proceedings. 22nd International Conference on*, April 2006, pp. 8–8.

[22] B. Hasani and M. H. Mahoor, "Facial expression recognition using enhanced deep 3D convolutional neural networks," in *CVPRW*, July 2017, pp. 2278–2288.

[23] A. Krizhevsky and G. Hinton, "Learning multiple layers of features from tiny images," *Master's thesis, Department of Computer Science, University of Toronto*, 2009.

[24] Y. Netzer, T. Wang, A. Coates, A. Bissacco, B. Wu, and A. Y. Ng, "Reading digits in natural images with unsupervised feature learning," in *NIPS Workshop on Deep Learning and Unsupervised Feature Learning 2011*, 2011.

[25] P. Viola and M. J. Jones, "Robust real-time face detection," *Int. J. Comput. Vision*, vol. 57, no. 2, pp. 137–154, May 2004.

[26] W. Ouyang and X. Wang, "Joint deep learning for pedestrian detection," in *ICCV*, Dec 2013, pp. 2056–2063.

[27] C. Szegedy, W. Liu, Y. Jia, P. Sermanet, S. Reed, D. Anguelov, D. Erhan, V. Vanhoucke, and A. Rabinovich, "Going deeper with convolutions," in *CVPR*, June 2015, pp. 1–9.

[28] C. Szegedy, V. Vanhoucke, S. Ioffe, J. Shlens, and Z. Wojna, "Rethinking the inception architecture for computer vision," *CoRR*, vol. abs/1512.00567, 2015.

[29] C. Szegedy, S. Ioffe, V. Vanhoucke, and A. A. Alemi, "Inception-v4, inception-resnet and the impact of residual connections on learning," in *AAAI*, 2017.

[30] A. G. Howard, M. Zhu, B. Chen, D. Kalenichenko, W. Wang, T. Weyand, M. Andreetto, and H. Adam, "Mobilenets: Efficient convolutional neural networks for mobile vision applications," *CoRR*, vol. abs/1704.04861, 2017.

[31] K. He, X. Zhang, S. Ren, and J. Sun, "Deep residual learning for image recognition," in *CVPR*, June 2016, pp. 770–778.

[32] K. He, Z. Xiangyu, S. Ren, and J. Sun, "Identity mappings in deep residual networks," in *ECCV 2016*, B. Leibe, J. Matas, N. Sebe, and M. Welling, Eds. Cham: Springer International Publishing, 2016, pp. 630–645.

[33] K. Simonyan and A. Zisserman, "Very deep convolutional networks for large-scale image recognition," *CoRR*, vol. abs/1409.1556, 2014.